\documentclass[runningheads]{llncs}
\usepackage[T1]{fontenc}
\usepackage{graphicx}
\usepackage{booktabs}

\usepackage{graphicx}
\usepackage{amsfonts,amsmath}
\usepackage{algorithm,algpseudocode}
\usepackage{mathrsfs}
\usepackage{nicefrac}
\usepackage{subcaption}
\usepackage{booktabs}
\usepackage{multirow}
\usepackage{xcolor}
\usepackage{comment}

\usepackage{hyperref}

\usepackage[misc]{ifsym}

\usepackage{mwe}

\usepackage[misc]{ifsym}

\begin{document}

\title{DWFL: Enhancing Federated Learning through Dynamic Weighted Averaging} 



\author{Prakash Chourasia\inst{1} \and
Tamkanat E Ali\inst{2}\and
Sarwan Ali \inst{1}
\and Murray Pattersn \inst{1}
}
\authorrunning{P. Chourasia et al.}
%
\institute{Georgia State University, Atlanta GA 30303, USA \and 
Lahore University of Management Science, Lahore, Pakistan
\email{pchourasia1@student.gsu.edu, 20100159@lums.edu.pk, sali85@student.gsu.edu, mpatterson30@gsu.edu}}



\raggedbottom
\maketitle

\begin{abstract}




Federated Learning (FL) is a distributed learning technique that maintains data privacy by providing a decentralized training method for machine learning models using distributed big data. This promising Federated Learning approach has also gained popularity in bioinformatics, where the privacy of biomedical data holds immense importance, especially when patient data is involved. Despite the successful implementation of Federated learning in biological sequence analysis, rigorous consideration is still required to improve accuracy in a way that data privacy should not be compromised. Additionally, the optimal integration of federated learning, especially in protein sequence analysis, has not been fully explored. We propose a deep feed-forward neural network-based enhanced federated learning method for protein sequence classification to overcome these challenges. Our method introduces novel enhancements to improve classification accuracy. We introduce dynamic weighted federated learning (DWFL) which is a federated learning-based approach, where local model weights are adjusted using weighted averaging based on their performance metrics. By assigning higher weights to well-performing models, we aim to create a more potent initial global model for the federated learning process, leading to improved accuracy. We conduct experiments using real-world protein sequence datasets to assess the effectiveness of DWFL. The results obtained using our proposed approach demonstrate significant improvements in model accuracy, making federated learning a preferred, more robust, and privacy-preserving approach for collaborative machine-learning tasks.


\end{abstract}

\section{Introduction}

The huge biological data generation like protein sequences, calls for advanced computation techniques. Several machine learning and deep learning-based methods have been used for protein classifications including support vector machine~\cite{cao2014protein}, convolution neural networks (CNN)~\cite{manavi2023cnn}, long short-term memory (LSTM)~\cite{ahsan2022classification}, and gated recurrent units (GRU)~\cite{le2019computational}. Due to the heavy reliance on centralized data processing, these traditional Machine Learning and Deep Learning techniques pose challenges related to privacy, computational efficiency, central data repository, scalability, and latency~\cite{paleyes2022challenges}.
Distributed machine learning approaches optimized their learning objectives to overcome these issues~\cite{nilsson2018performance} but they pose communication challenges. Also, the training data per local model in a distributed setting is assumed to be much larger~\cite{mcmahan2017communication}. 

Federated Learning was first proposed by Google~\cite{mcmahan2017communication} as a novel approach to train Machine Learning models across different devices while preserving data privacy by conducting local computations on each device avoiding the need to share local data~\cite{beltrán2023fedstellar}. Federated Learning (FL), an emerging and promising Artificial Intelligence paradigm ~\cite{mcmahan2023communicationefficient}, has the potential to overcome the limitations of distributed computing while keeping data privacy. 
The FL model consists of three main steps: The first step is global model initialization the weights are then sent to the local clients participating in FL. In the second step, the local models are initialized using the weights from the previous step and trained using the local data. The updated local model parameters (weights) are sent to the central server to transmit the learning from local data (in an abstract manner). In the third step, the central server aggregates the local model parameters received by the central server and generates an updated global model. These three steps are repeated iteratively for continuous updates.  



There have been recent developments to optimize FL-based models like Federated Averaging (FedAvg)~\cite{mcmahan2017communication}, and Federated Stochastic Gradient Descent (FedSGD)~\cite{saha2021federated}. FedSGD collects parameters learned from a single communication update from a central server~\cite{saha2021federated}. Despite its promise, federated learning faces various challenges. Like, even the involvement of one poor-performing local model can negatively impact the global model. The traditional methods have failed to dynamically aggregate local models according to the individual performance of local clients with local data~\cite{tu2021feddl}. 


This research aims to develop an improved federated learning-based architecture that can perform biological sequence classification more accurately while penalizing the poor-performing model while aggregating. We enhance the federated learning process while considering the local model performance and local data quality. We then classify protein sequences using this improved proposed dynamic weighted federated learning (DWFL) mechanism. 
The Global Model generated based on DW-FedAvg is represented by the expression $G=\dfrac{1}{N}\sum_{i=1}^N \beta_i \cdot L_i$. Here $G$ is the global model created at time $t$, $N$ is the total number of local models, $L_i$ is the $i$th local model received at the global server at time $t-1$ and $\beta_i$ is the dynamic weight associated with the $i$th local model. The proposed mechanism proves to be successful in improving the predictive performance. Our contributions can be summarized as follows:

\begin{enumerate}

    \item 
    To the best of our knowledge, our work is the first one to present a Dynamic Weighed Federated Learning (DWFL) framework for protein sequence classification which is capable of constructing a highly accurate global classifier without accessing the local data while classifying protein sequence.
    \item The proposed federated learning involving Dynamic Weighted averaging strategy and Feed-Forward Neural Network is capable of constructing a highly accurate global classifier without accessing the local data while classifying protein sequence.
    \item We performed experiments with baseline methods to compare with our proposed dynamic weighted federated learning DWFL approach and assess their effectiveness in real-world federated learning scenarios involving protein classification.
    \item We use t-SNE for the visualization in order to evaluate the proposed model and other baseline methods.

\end{enumerate}



The rest of the paper is organized as follows: Section~\ref{sec_rw} provides a detailed literature review. Section~\ref{sec_PA} discusses the proposed method. Section~\ref{sec_ES} describes the experimental setup, datasets, evaluation metrics, and data visualization. Experimental results and performance analysis are reported in Section~\ref{sec_ER}. Finally, Section~\ref{sec_conclusion} concludes the paper.


\section{Related Work}
\label{sec_rw}

Several machine learning and deep learning-based methods have been used for protein sequence classifications. CNN-based amino acid representation learning~\cite{zhang2020protein} was applied taking into consideration the information of amino acid's location to explore the performances of annotated protein families. A deep CNN~\cite{szalkai2018near} was trained to classify protein sequences. Deep-gated recurrent units architecture has been used widely in computational biology problems involving protein sequences~\cite{le2019fertility}. GRU has also been used for the computational identification of vesicular transport proteins from sequences~\cite{le2019computational}. In a recent work~\cite{ahsan2022classification}, a novel LSTM (Long Short-Term Memory) deep learning neural network algorithm was used to classify imbalanced influenza virus A subtypes based on their amino acid sequences~\cite{ahsan2022classification}. 

Recently novel deep learning, kernel function~\cite{ali2024mik}, and machine learning-based models have been presented to classify proteins~\cite{seo2018deepfam,murad2023advancing,ali2024deeppwm,ali2024molecular,ali2024nearest,murad2024dance,ali2024compression,ali2024position}. However, these models do not deal with hierarchical protein classification~\cite{ali2024gaussian,ali2024elliptic}. To overcome this limitation DeepHiFam~\cite{sandaruwan2021improved} was introduced. Researchers in a study~\cite{choi2017gram}, proved that patient models can be improved by incorporating clinical concept hierarchies. The Poincaré embeddings~\cite{nickel2017poincare} accurately capture these hierarchical relations and hence are very suitable for healthcare applications.
Poincar\'e encoding~\cite{nickel2017poincare,ali2024preserving} is suitable for hierarchical data as it embeds labels in a hyperbolic space where distances between points are measured using the Poincar\'e distance. Protein Encoder~\cite{manzoor2023protein} is an unsupervised deep learning method that uses autoencoder with feature selection techniques. The autoencoder learns the non-linear relationship between features and finds an accurate representation~\cite{manzoor2023protein}. 

Despite the success of these ML methods, there are certain challenges related to privacy and dependence on centralized data processing~\cite{paleyes2022challenges}.


A Federated learning-based framework offers immense potential for preserving privacy in bioinformatics-related tasks. Evident by researchers while using the FL method to diagnose the type of skin disease by classifying dermoscopy images~\cite{hashmani2021adaptive} on medical image datasets. In another research deep neural network-based FL architecture was used to analyze ultrasound images to detect whether thyroid nodules were normal or dangerous~\cite{lee2021federated}.
Using chest X-ray images~\cite{diaz2023study} researchers using a federated learning strategy tried to predict whether the patient had pneumonia or not while preserving privacy.
The Federated Learning frameworks vary in the adaptive aggregation methods employed. For example, FedAvg~\cite{raimondi2023genome} stands for Federated Averaging it is a translation of Stochastic Gradient Descent (SGD) to Federated Learning, in each Federated Learning round the local client models send updated parameters to the Central Server which calculates their weighted average to create a new global model which is shared back to the local clients when the next Federated Learning round begins. FedAvg might suffer from slow convergence in certain data and class unbalancement situations~\cite{reddi2020adaptive}. FedAvgM an extension of FedAvg has an additional momentum term in the updates received from the local clients, this results in improved convergence speed and reduces the effect of noisy updates~\cite{hsu2019measuring}. 
Federated Learning has also been applied to biological sequence classification tasks~\cite{chourasia2023efficient}.In~\cite{chen2023privacy}, a Federated Learning model is used for DNA motif discovery tasks. This method solves privacy issues and data silo problems. FedAvg is used in Parkinson’s disease prediction~\cite{danek2024federated}, rare cancer classification~\cite{d2024mosaic}, anti-cancer drug senstivity prediction~\cite{xu2023predicting} etc. In~\cite{chourasia2023efficient}, an efficient method was proposed for the classification of SARS-CoV-2 Spike Sequences Using Federated Learning. However, these approaches do not consider the local model performance while aggregating weights. 


\section{Proposed Approach}\label{sec_PA}
We describe the proposed dynamic weighted federated-learning-based (DWFL) method for sequence classification in this section. We start by outlining the overall architecture design of the suggested model, then discuss in detail the algorithm and workflow.

\subsection{Architecture}

The architecture for the proposed DWFL approach comprised three building blocks: a) embedding Generation b) Dataset Division and c) Global Model and Local Model Aggregation. 
The first step is to generate numerical Embedding for a given biological sequence using One-Hot Encoding (OHE)~\cite{kuzmin2020machine}. In the second step, we divide the dataset to allocate it to local models. The data is divided into Global Test and Train Dataset. The training dataset is further subdivided into Local model training data and Global Model Train Dataset. The Local Model Train Dataset is further divided into equal sets for each local Model (see the remark~\ref{div_rason}). In our experiments, we took $6$ local models. 

\begin{remark}
\label{div_rason}
In a real-world scenario, the second step to divide the dataset is not required since all local models will be trained on a local dataset that is private to the local clients. Here we are segregating data to demonstrate and perform the experiments.
\end{remark}

The final component is model Aggregation where the the Global model is updated with the aggregated weights from the local model. While aggregation we count the dynamic weights from local models to make informed aggregation. 
The idea is to include the dynamic weight while aggregating the weights from local models. 
This extracts useful information from the local dataset owned/produced by local clients. By not requesting that the local clients transfer the actual data to the central server, data privacy is maintained. Local models are trained using a local dataset and output weights that represent the retrieved data. To initialize a global model with these incoming weights after aggregation, we use the suggested dynamic weighted federated learning DWFL method (described in subsequent paragraphs). For SARS-CoV-2 spike protein sequence classification, our suggested DWFL architecture consists of feed-forward neural networks for local and global settings. On various local datasets, we train several local models, and their weights are then combined while taking the dynamic weight into account to aggregate it into a global model.



\subsection{Algorithm and Workflow}
\begin{figure}[h!]
    \centering
    \includegraphics[scale=0.18]{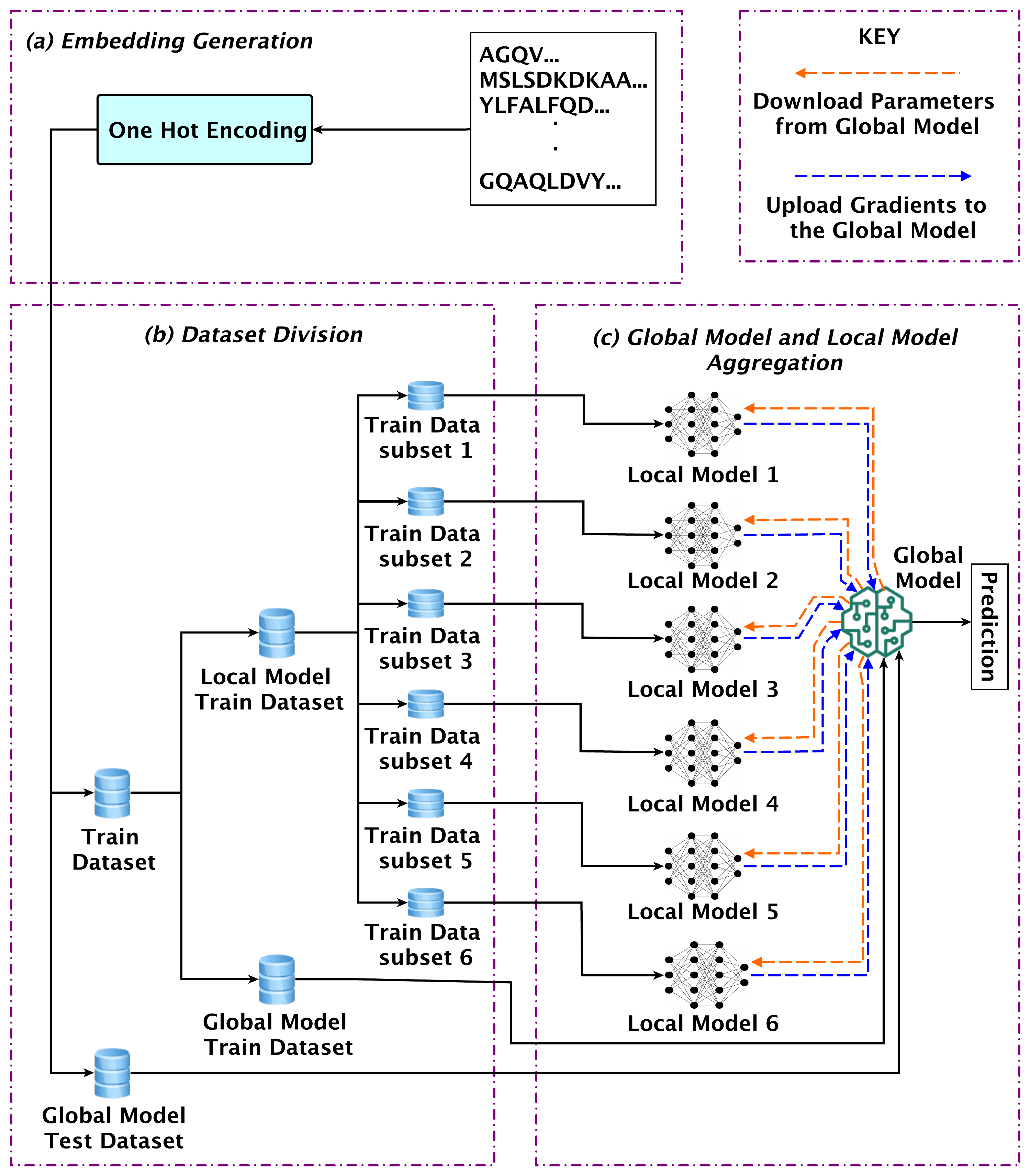}
\caption{The Proposed Federated Learning Approach.}
\label{Architecture_DWFL}
\end{figure}

The proposed DWFL is an FL-based approach having three components including local models, global models, and iterative weight updates. 
The Algorithm~\ref{algo_fedrated} shows the pseudocode of our proposed dynamic weighted federated learning (DWFL) model, whereas the Figure~\ref{Architecture_DWFL} shows the workflow. As we can see in the workflow we have divided it into 3 sections including a) embedding generation, b) dataset division, and c) global model and local model aggregation. The algorithm and workflow are discussed in detail below:

For given protein sequences and lineages (target labels), we generate the numerical vectors (feature vectors) by employing a one-hot encoding~\cite{kuzmin2020machine} on both sequences and labels as shown in line 1 and line 2 of Algorithm~\ref{algo_fedrated} also shown in Figure~\ref{Architecture_DWFL}-a. The resultant vectors and labels are then divided into training and test sets  $X_{tr}$, $y_{tr}, X_{t}$, $y_{t}$ as shown in line 3 of Algorithm~\ref{algo_fedrated} also can be seen in Firgure~\ref{Architecture_DWFL}-b. The training set $X_{tr}$, $y_{tr}$ are further divided into Local Model training dataset and Global Model training dataset as shown in line 4 of Algorithm~\ref{algo_fedrated} and  Firgure~\ref{Architecture_DWFL}-b. In the final data split the Local training dataset $X_{ltr}$ is segregated based on $6$ different and equal datasets for local models, as shown in lines 6 and 7 of the Algorithm~\ref{algo_fedrated} (also the final step of Figure~\ref{Architecture_DWFL}-b. Each local training dataset has feature vectors $X_{li}^{ltr}$ as the origin for the spike sequences and labels $y_{li}^{ltr}$. 

\begin{algorithm}[H]
\scriptsize
    \caption{DWFL Model Pseudo Code}
    \label{algo_fedrated}
	\begin{algorithmic}[1]
	\Statex \textbf{Input:} Feature vectors $X$, Labels $y$, and $localModelList$ 
	\Statex \textbf{Output:} Host predictions $y$
	\State $X$ = oneHotEncoding($X$)
        \State $y$ = oneHotEncoding($y$)
        \State $X_{tr}$, $y_{tr}, X_{t}$, $y_{t}$ = testTrainSplit($X$, $y$)
        \State $X_{ltr}$, $y_{ltr}, X_{gtr}$, $y_{gtr}$ = localGlobalTrainSplit($X_{tr}$, $y_{tr}$) 
	\For{$i$ in $localModelList$}
	    \State $X_{li}^{ltr}$ = modelWiseSeq($X_{ltr}$) \Comment{seq for local model i}
	    \State $y_{li}^{ltr}$ = modelWiseLabels($y_{ltr}$) \Comment{labels}
            \State lm = newLocalModel()         \Comment{Local model}
       	\State $lm_{i}$ = lm.LocalModelTraining($X_{li}^{ltr},y_{li}^{ltr}$)
            \State $valAcc\_lm_{i}$ = $lm_{i}$.getValAcc(split = 0.1)
            \State $valAccList$.append($valAcc\_lm_{i}$) \Comment{val acc list}
        \EndFor{}
        \State $totalValAccuracy$ = sum($valAccList$)
        \For{$i$ in $localModelList$}
            \State $acc = lm_{i}[valAcc]$
            \State $dynanmicWt[i]$ = $\frac{acc}{totalValAccuracy}$
            \State $wList.append(lm_{i}[w_i] \times dynanmicWt[i])$
        \EndFor{}
        
	\State $avgWeights$ = Aggregation($wList$) \Comment{Avg/min/max}
	\State $gm$ = globalModelWeightsUpdate($avgWeights$) 
        \State $gm$.GlobalModelTraining($X_{gtr}, y_{gtr}$) 
	\State $pred$ = $gm$.Predict($X_{t}, y_{t})$ \Comment{testing Global model}
        \State return($pred$)
    \end{algorithmic}
\end{algorithm}
\graphicspath{ {./Figures/} }

After the data split, the training takes place where each local model is trained with their respective local dataset as shown in line 9 of Algorithm~\ref{algo_fedrated} and Figure~\ref{Architecture_DWFL}-c. Now we collect the validation accuracy of each model with 10\% data split and create a list of all validation accuracies for local models as can be seen in lines 10 and 11 of Algorithm~\ref{algo_fedrated}. We compute the dynamic weight component for each model using these cumulative summed-up validation accuracies for all models. We normalize the accuracy of each model by dividing it by the total summed-up accuracy for all other local models to get the $dynanmicWt[i]$ dynamic weight for local model i, which can be seen in lines 13,14, and 15 of the Algorithm~\ref{algo_fedrated}.  We infuse this dynamic weight component to adjust the weight of the respective local model and create a list of all dynamically adjusted weights of local models as shown in line 16 of Algorithm~\ref{algo_fedrated}. These weights are aggregated to provide the $avgWeights$ which are used to update the weights for the global model (line 18 of the  Algorithm~\ref{algo_fedrated}). In the final step Global model is trained using the global training dataset $X_{gtr}$ and $y_{gtr}$. This can be repeated iteratively any arbitrary number of n times to keep updating the weights (in our experiments we use n=1). Finally, after the global model is trained we use the test data set $X_t$ and $y_t$ to predict and evaluate the model using different evaluation metrics.

\subsection{Neural Network Model Architecture}
\label{sec_architecture_appendix}

The model’s architecture we use in our proposed DWFL is shown in Table~\ref{tbl_nn_params}. It is composed of a sequence of Dropout layers (value = 0.5), Batch Normalization layers, and Dense layers using L1 regularization. 5 times in the same order, this trio of layers is combined. An output-dense layer comes next. In this instance, ADAM is utilized as an optimizer, while softmax is the activation function. 

\begin{table}[h!]
  \centering
  \resizebox{0.65\textwidth}{!}{
    \begin{tabular}{p{2.5cm}ccp{1.5cm}}
    \toprule
        Layer (type) & Output Shape  & Trainable Parameters \\
        \midrule \midrule
        Input Layer & \begin{tabular}{c} Output : (None, 26754)\end{tabular} & - \\
        \midrule
        Dense Layer & \begin{tabular}{c} Output : (None, 512)\end{tabular} & 13,698,560 \\
        \midrule
        Batch Normalization & \begin{tabular}{c} Output : (None, 512)\end{tabular}  & 2048 \\
        \midrule
        Dropout Layer & \begin{tabular}{c} Output : (None, 512)\end{tabular}  & - \\
        \midrule
        Dense Layer & \begin{tabular}{c} Output : (None, 256)\end{tabular}  & 131328 \\
        \midrule
        Batch Normalization & \begin{tabular}{c} Output : (None, 256)\end{tabular}  & 1024 \\
        \midrule
        Dropout Layer & \begin{tabular}{c} Output : (None, 256)\end{tabular}  & - \\
        \midrule
        Dense Layer & \begin{tabular}{c} Output : (None, 128)\end{tabular}  & 32896 \\
        \midrule
        Batch Normalization & \begin{tabular}{c} Output : (None, 128)\end{tabular}  & 512 \\
        \midrule
        Dropout Layer & \begin{tabular}{c} Output : (None, 128)\end{tabular}  & - \\
        \midrule
        Dense Layer 1 & \begin{tabular}{c} Output : (None, 64)\end{tabular} & 8256 \\
        \midrule
        Batch Normalization & \begin{tabular}{c} Output : (None, 64)\end{tabular}  & 256 \\
        \midrule
        Dropout Layer & \begin{tabular}{c} Output : (None, 64)\end{tabular}  & - \\
        \midrule
        Dense Layer 1 & \begin{tabular}{c} Output : (None, 32)\end{tabular} & 2080 \\
        \midrule
        Batch Normalization & \begin{tabular}{c} Output : (None, 32)\end{tabular}  & 128 \\
        \midrule
        Dropout Layer & \begin{tabular}{c} Output : (None, 32)\end{tabular}  & - \\
         \midrule
        Dense Layer 1 & \begin{tabular}{c} Output : (None, 22)\end{tabular} & 726 
        \\
        \midrule
        \midrule
        Total & \_  & 13,875,830 \\
        \bottomrule
    \end{tabular}
    }
  \caption{Neural Network model architecture.}
  \label{tbl_nn_params}
\end{table}

\section{Experimental Setup}\label{sec_ES}
The information regarding the classification models and the appropriate assessment criteria used to report the performance is highlighted in this section along with specifics of the datasets used to carry out the experiments. 
The Intel (R) Core i5 machine, which has a 2.40 GHz processor and $32$ GB of memory, is used for all experiments. The experiments are carried out using Python. 

\subsection{Dataset Statistics}


The Virus Pathogen Database~\footnote{\url{https://www.viprbrc.org/}} provided the protein $2576$ total sequences from several types of mammalian and avian coronaviruses (ViPR)~\cite{pickett2012vipr}. Different fasta files (one fasta file per host) were used to hold the amino acid sequences from each host.
Clustal Omega was used to perform multiple sequence alignments~\footnote{\url{https://www.ebi.ac.uk/Tools/msa/clustalo/}}. 
The dataset contains about $12$ host types of sequences that we gathered through the annotation of $2574$ complete protein sequences. The detail of each host, i.e., name (count/distribution), in the dataset, is as follows:   Bat (181), BelugaWhale (2), Camel (265), Cat (57), Chicken(309), Chimpanzee(3), Civet(5), Duck(3), Goose(2), Human(957), Pangolin(5 ), Swine(785). It is also shown in Table~\ref{tbl_dataset_statistics}. 

\begin{table}[h!]
    \centering
    \resizebox{0.5\textwidth}{!}{
    \begin{tabular}{cc|cc}
    \hline
        Host Name & \# of Sequences & Host Name & \# of Sequences \\
        \hline \hline
        Bat & 181 & BelugaWhale & 2     \\
        Camel & 265  & Cat &  57           \\
        Chicken &  309 & Chimpanzee &  3     \\
        Civet &  5 & Duck &  3           \\
        Goose &  2 & Human &  957        \\
        Pangolin &  5  & Swine &  785        \\
        \midrule
         -  & - & \textbf{Total} & \textbf{2574}\\
         \bottomrule
        \hline
    \end{tabular}
    }
    \caption{Statistics for Host dataset.}
    \label{tbl_dataset_statistics}
\end{table}

\subsection{Baseline Models}
We evaluate our model by comparing it with various state-of-the-art models belonging to different domains including Neural Networks models (Autoencoder + NTK, CNN, LSTM, GRU, Feed-Forward Neural Networks, and FedAvgWeight) and Hyperbolic embedding-based models (Poincaré Embedding).  

\subsubsection{Poincaré Embedding}
Poincaré Embedding~\cite{nickel2017poincare} uses the Poincar\'e distance formula 
$ d(a_1,a_2)=arcCosh \left(1+2\dfrac{\lVert a_1-a_2 \rVert^2}{(1-\lVert a_1 \rVert^2)(1-\lVert a_2 \rVert^2)}\right) $
to map the amino acid sequence into a hyperbolic space. 
Here $a_1$ and $a_2$ represent two points (amino acids) in p-dimensional unit ball and $\lVert.\rVert$ denotes Euclidean norm. The equation shows the distance changes smoothly concerning the location of $a_1$ and $a_2$ within the Poincaré ball.

\subsubsection{Autoencoder + NTK} 
This approach~\cite{nguyen2021benefits} involves Autoencoder which generates low dimensional embedding and Neural Tangent Kernel (NTK)~\cite{jacot2018neural} which is a kernel function that computes pairwise kernel values based on the geometry of the neural network’s decision boundary~\cite{chourasia2023empowering}. A 4-layer autoencoder is used where the loss function selected is Mean Squared Error and the optimizer is ADAM. It was run for 10 epochs. The autoencoder consists of an encoder and a decoder. The encoder contains dense layers and a dropout layer. The activation function used in the case of dense layers is LeakyReLU, additionally, batch normalizations are also applied. The decoder decreases the reconstruction loss and reconstructs the embedding. Finally, the embedding is computed by using Kernel PCA~\cite{hoffmann2007kernel} and the top principal components from the NTK kernel matrix.

\subsubsection{LSTM}
It is a type of RNN model and is known for its ability to model sequential data. In LSTM~\cite{hochreiter1997long}, information flow is controlled using the gating mechanism, which deals with the gradient vanishing problem and also captures long-range dependency. The sequences of layers in the neural network architecture are as follows first comes the embedding layer where embedding is set to 500, followed by 2 sets of LSTM layer and LeakyReLU layer in which both the LSTM layers contain 200 memory units, and for the two LeakyReLU layer the alpha is set to 0.05, then comes a dropout layer (value of 0.2), a Dense layer (dimensions = 500), another LeakyReLU layer, finally followed by the output layer the optimizer chosen in this case is ADAM~\cite{kingma2014adam} and Sigmoid activation.

\subsubsection{GRU}
GRU networks~\cite{cho2014properties} are also a type of Recurrent Neural Networks that are designed to learn long-term and short-term relationships within the sequence data~\cite{chung2014empirical}. The specifications of these layers are as follows embedding is set to 500 in the embedding layer, for the GRU layer 200 memory units are used, a LeakyReLU layer is included with alpha equal to 0.05, and the Dropout layer is added with the value set to 0.2. ADAM optimizer and the Sigmoid activation function are employed. 

\subsubsection{CNN}
Convolutional Neural Network (CNN) is a deep-learning technique where a series of convolutional layers are used to extract features from the input data~\cite{gunasekaran2021analysis}. The sequence of the layers is as follows a one-dimensional convolution layer, then a LeakyReLU layer, and a batch normalization layer. Followed by another set of these three layers. Each one-dimensional convolutional layer consists of 128 filters and a kernel of size 5. The size of the kernel plays a significant role~\cite{gunasekaran2021analysis}. The alpha in both the LeakyReLU layers is set to 0.05. Then comes the Max pooling layer (pool size = 2), where the dimensions of extracted features are reduced, this is followed by three more layers namely dense layer (dimension = 500), LeakyReLU layer (alpha = 0.05), and output dense layer. We use the ADAM optimizer and the sigmoid activation function. 

\subsubsection{Centralized Feed Forward NN}
The feed-forward neural network (FNN)~\cite{glorot2010understanding} is a multilayer network in a centralized setting. Having the same architecture as our global model. The model’s architecture we use is shown in Table~\ref{tbl_nn_params}. It consists of a series of Dense layers with L1 regularization, Batch Normalization layers, and Dropout layers (value = 0.5). This combination of three layers is repeated 5 times in the same order. Followed by an output dense layer. The activation function used in this case is softmax and ADAM is used as a optimizer.   

\subsubsection{FedAvgWeight}
The Federated learning-based approach FedAvgWeight~\cite{chourasia2023empowering} trains the feed-forward neural network locally and sends the updated weights to the central server by taking average weight values from the local models to aggregate the learning from local data. It provides data privacy and reduces computational costs and latency issues since data is not transmitted to a central server or anywhere. FedMinWeight and FedMaxWeight are the variations that we employed in our experiments where minimum and maximum weights are chosen to be sent to a central server.  Thus a federated learning-based model provides data security since there is no need to share the data (data does not move out of premises, and only parameters or weights are pushed back). 

\subsection{Evaluation Metrics}
We use the Naive Bayes (NB), Multilayer Perceptron (MLP), k-nearest Neighbour (k-NN) (where $k = 3$), Random Forest (RF), Logistic Regression (LR), and Decision Tree (DT) ML models for classification tasks. To preserve the original data distribution, the data for each classification task is divided into 70-30\% train-test sets using stratified sampling. To produce more consistent findings, our studies also average the performance outcomes of $5$ runs for each combination of dataset and classifier. Accuracy, precision, recall, weighted F1, F1 macro, and ROC AUC macro were the performance metrics we used to assess the classifiers. Because we are performing multi-class classification in some circumstances, we used the one-vs-rest method to calculate the ROC AUC value. Additionally, the goal of presenting a variety of metrics is to gain additional knowledge about the performance of the classifiers, particularly in the case of class imbalances where reporting accuracy alone is insufficient to understand performance. 

\subsection{Data Visualization}

Visualizing the data is a very important step when it comes to gaining clarity of the nature of the data, which in turn helps in performing accurate analysis of the results. Mostly the biomedical field deals with high-dimensional data which is not that easy to visualize in its original form. In our case, we have a high dimensional SARS-CoV-2 spike protein sequence dataset, so we need to reduce the dimensions of the data to visualize the data in two dimensions. 

\begin{figure*}[h!]
  \centering
  \begin{subfigure}{.33\textwidth}
  \centering
   \includegraphics[scale=0.07]{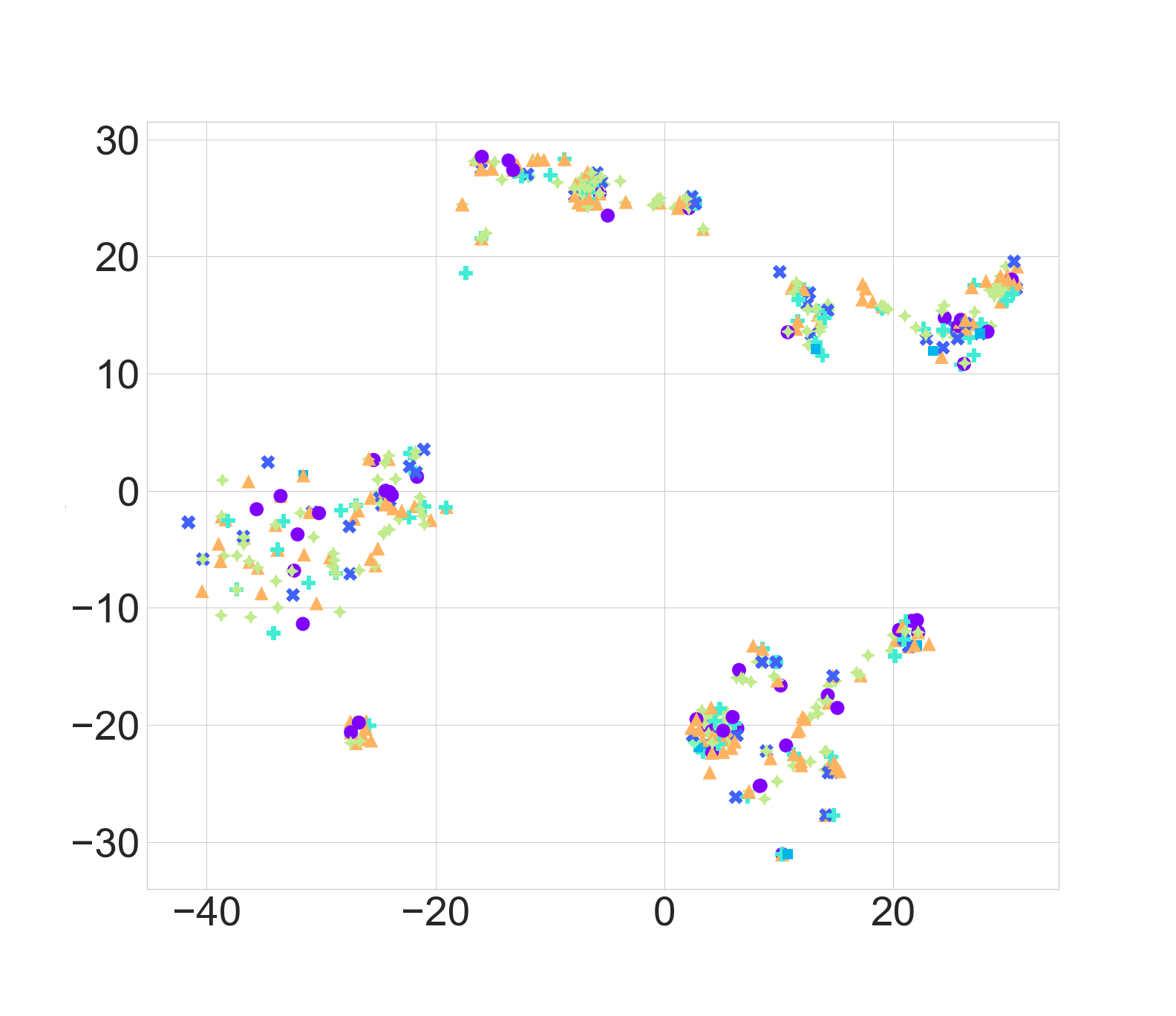}
    \caption{Global Model}
    \label{fig_tsne_GM}
  \end{subfigure}%
  \begin{subfigure}{.33\textwidth}
  \centering
  \includegraphics[scale=0.07]{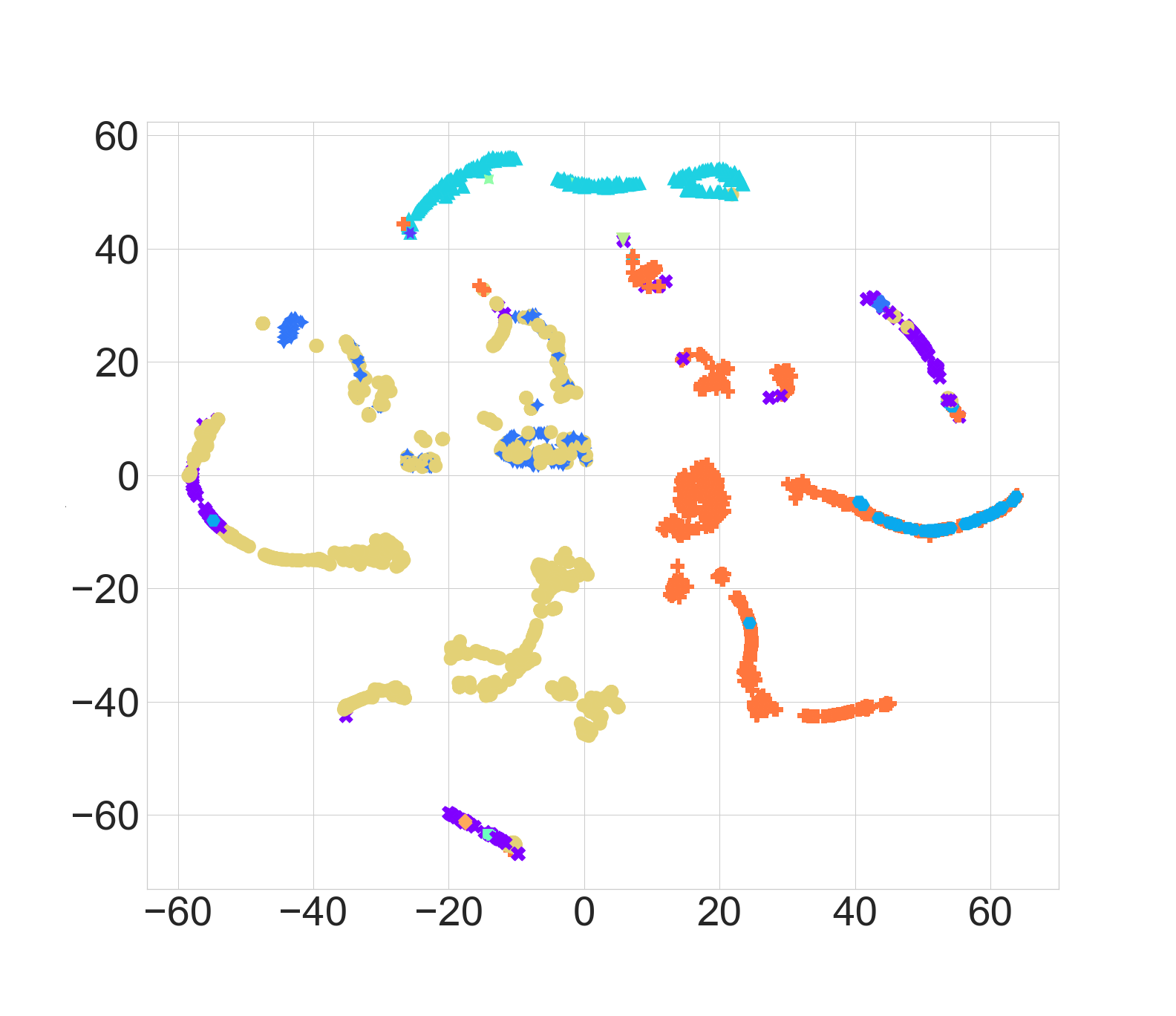}
    \caption{AE + NTK}
    \label{fig_tsne_Auto_NTK}  
  \end{subfigure}%
  \begin{subfigure}{.33\textwidth}
  \centering
  \includegraphics[scale=0.07]{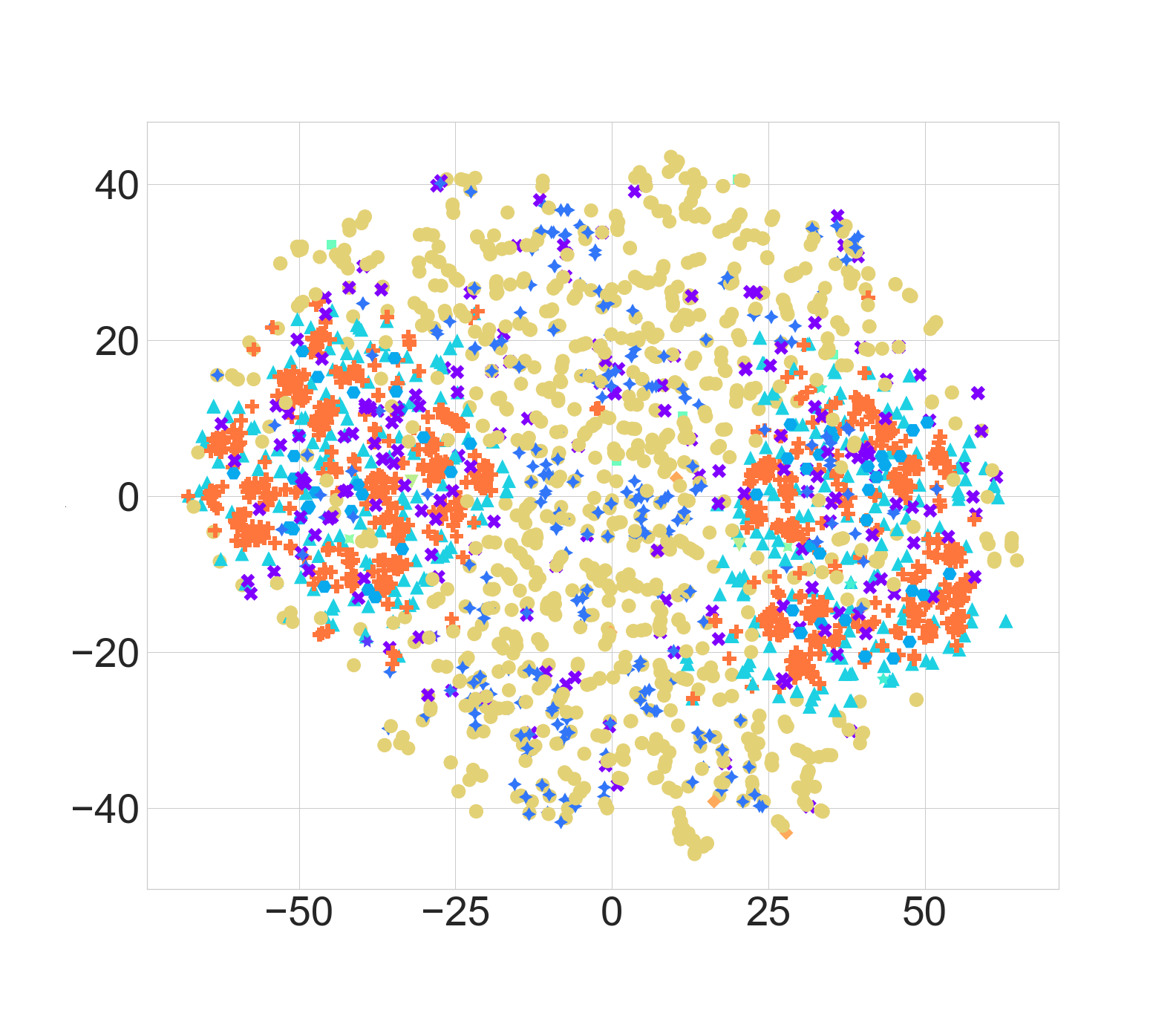}
    \caption{Poincare}
    \label{fig_tsne_Poin}
  \end{subfigure}%
  \\
  \begin{subfigure}{.99\textwidth}
    \centering
    \includegraphics[scale=0.32]{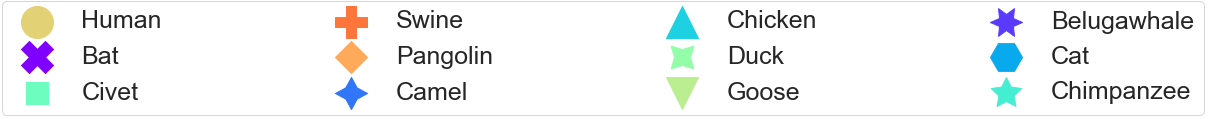}
  \end{subfigure}
  \caption{t-SNE Plot (Global Model and Baselines). The figure is best seen in color.}
  \label{fig_tsne}
\end{figure*}

\begin{figure*}[h!]
  \begin{subfigure}[t]{0.33\textwidth}
  \centering
    \includegraphics[scale=0.08]{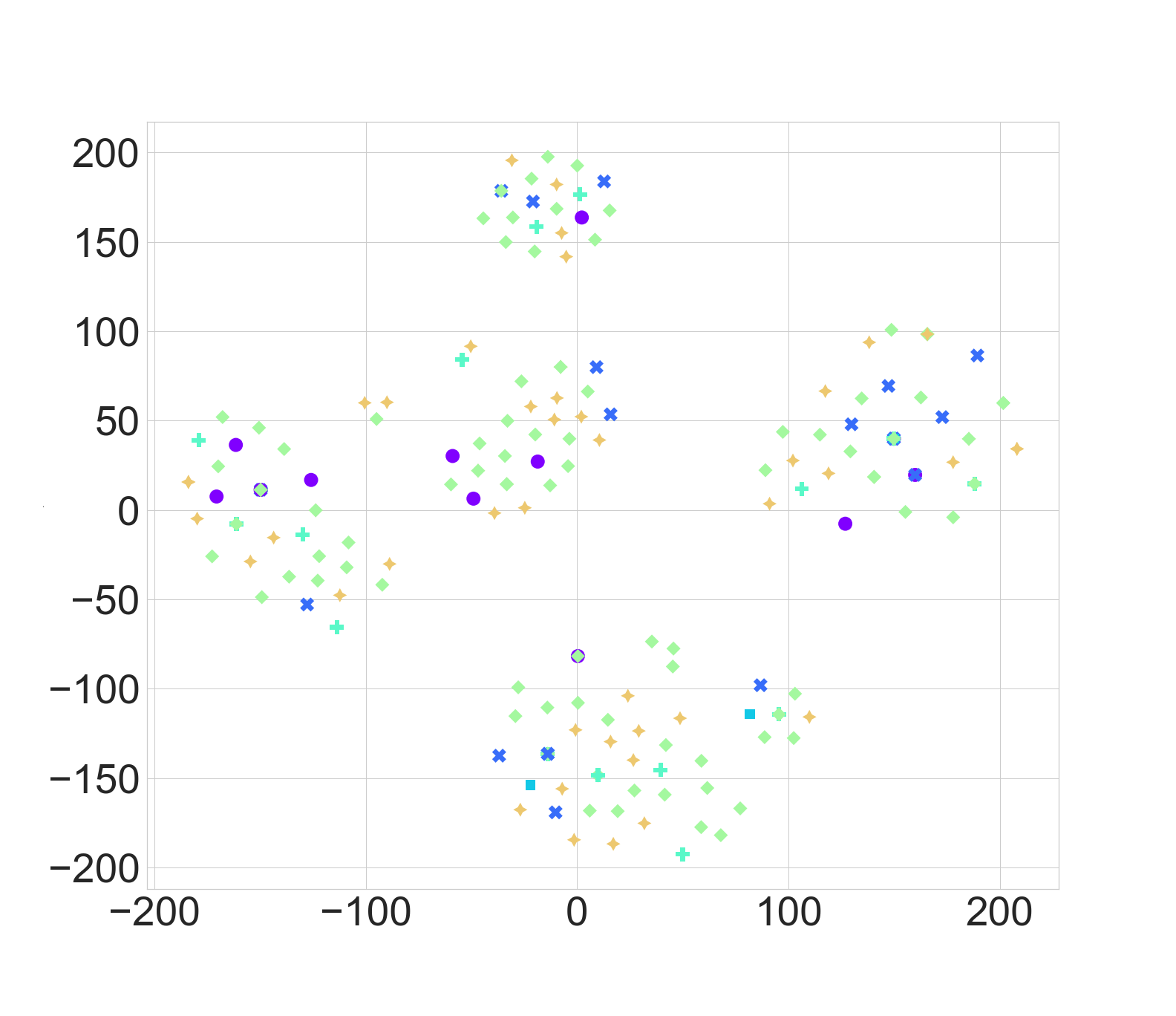}
    \caption{Local Model 1}
    \label{fig_tsne_LM1}
  \end{subfigure}%
  \begin{subfigure}[t]{0.33\textwidth}
  \centering
    \includegraphics[scale=0.08]{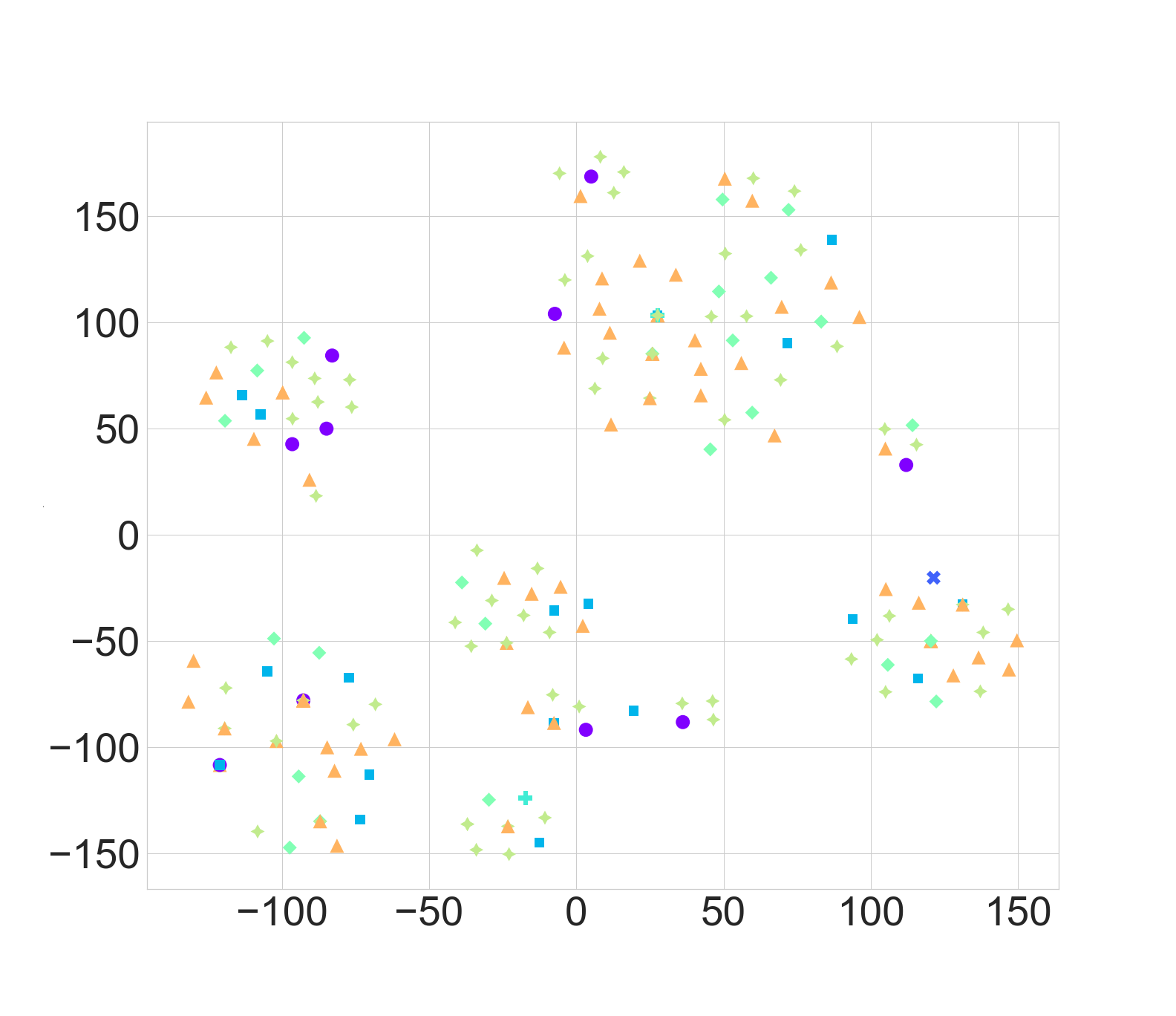}
    \caption{Local Model 2}
    \label{fig_tsne_LM2}
  \end{subfigure}%
  \begin{subfigure}[t]{0.33\textwidth}
  \centering
    \includegraphics[scale=0.08]{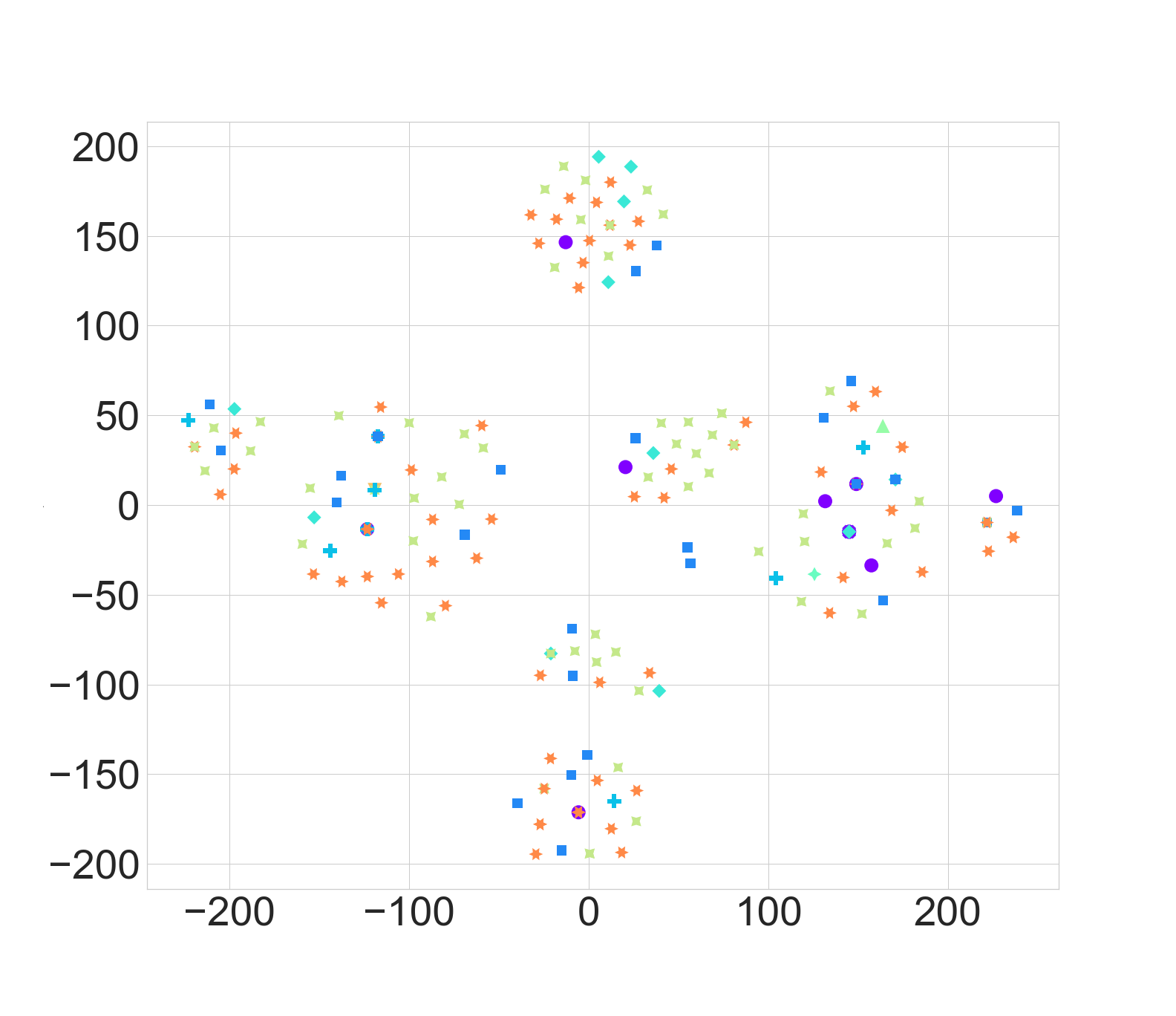}
    \caption{Local Model 3}
    \label{fig_tsne_LM3}
  \end{subfigure}%
  \\
  \begin{subfigure}[t]{0.33\textwidth}
  \centering
    \includegraphics[scale=0.08]{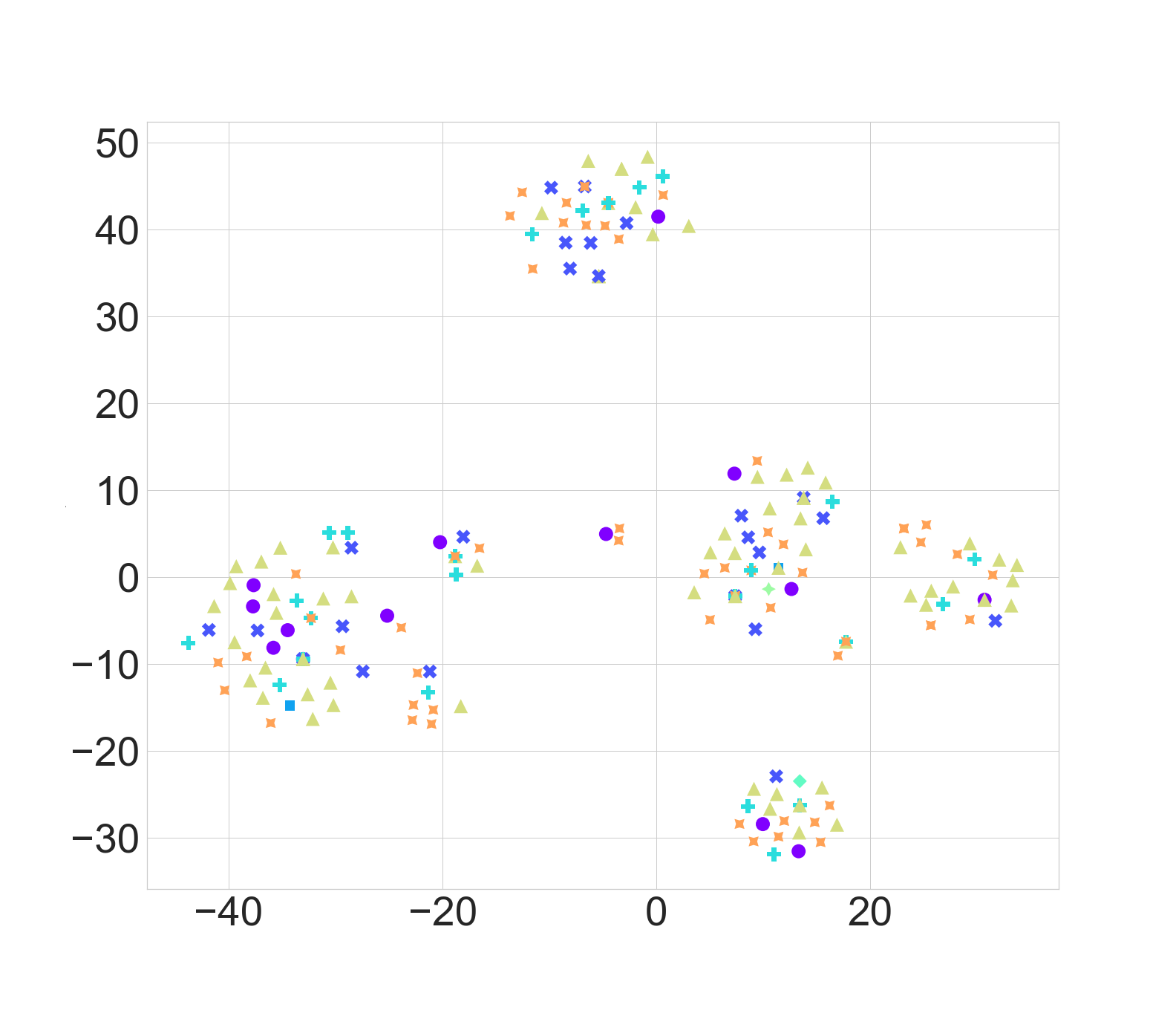}
    \caption{Local Model 4}
    \label{fig_tsne_LM4}
  \end{subfigure}%
  \begin{subfigure}[t]{0.33\textwidth}
  \centering
    \includegraphics[scale=0.08]{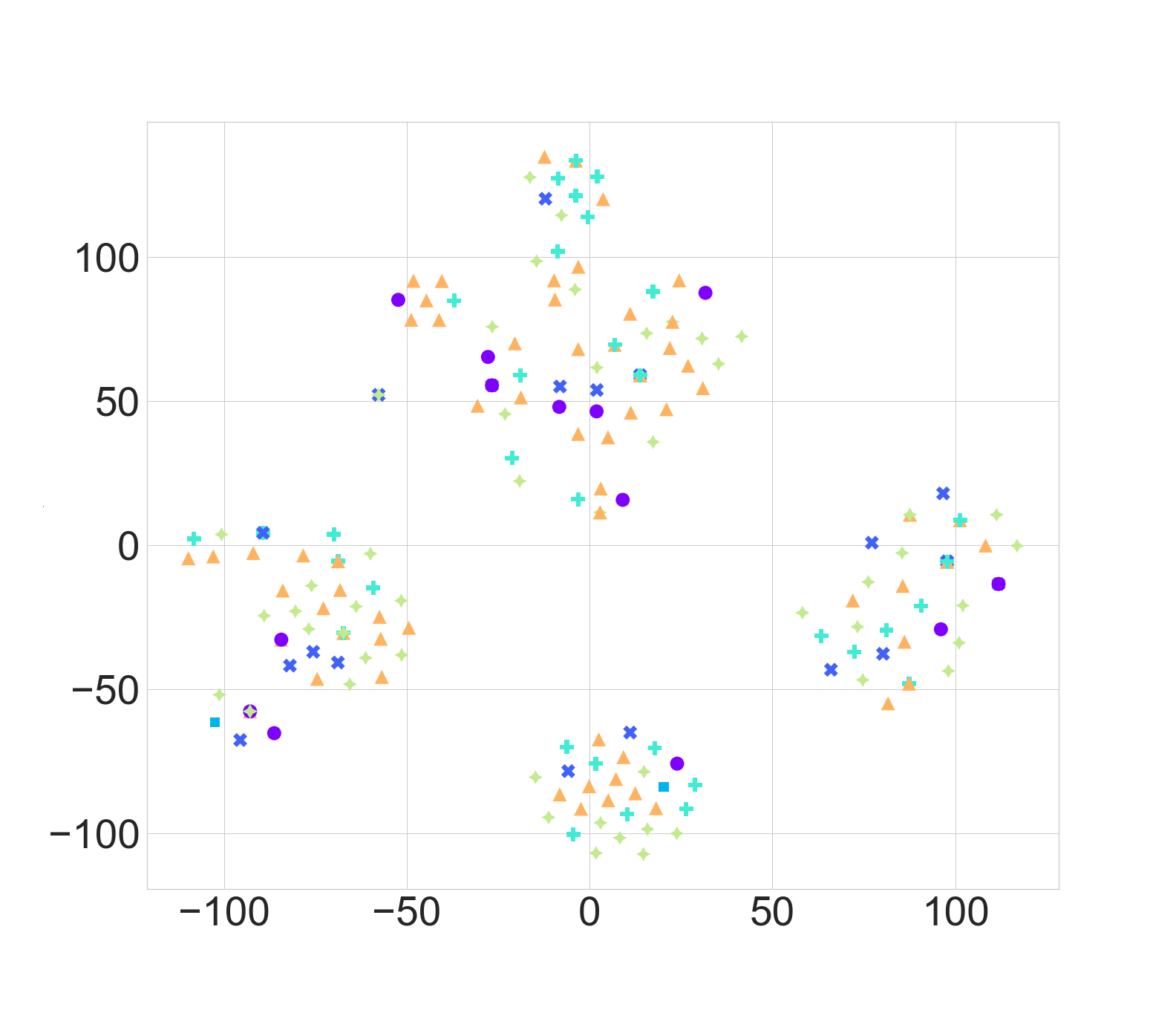}
    \caption{Local Model 5}
    \label{fig_tsne_LM5}
  \end{subfigure}%
  \begin{subfigure}[t]{0.33\textwidth}
  \centering
    \includegraphics[scale=0.08]{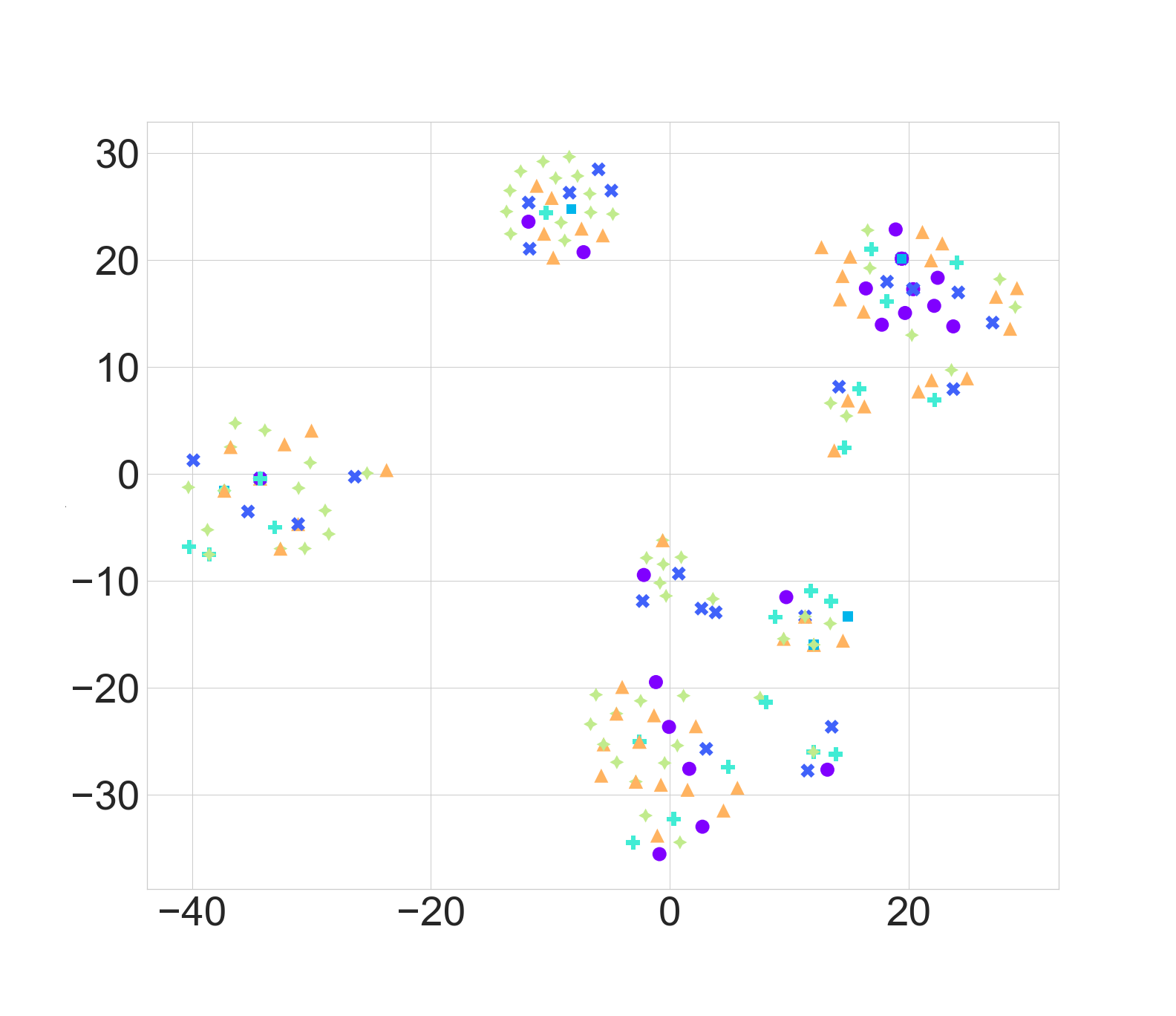}
    \caption{Local Model 6}
    \label{fig_tsne_LM6}
  \end{subfigure}%
  \\
  \begin{subfigure}[t]{0.99\textwidth}
  \centering
    \includegraphics[scale=0.3]{Figures/tSne/Legend.png}
  \end{subfigure}
  \caption{t-SNE plots for Local and Global Models along with baselines. The figure is best seen in color.}
  \label{fig_tsne_localmodels}
\end{figure*}

For visualization, we use t-SNE~\cite{van2008visualizing} a popular dimensionality reduction technique in biological data which is used for visualizing high-dimensional data in a lower-dimensional. space~\cite{butt2023ensemble}. 
Figure~\ref{fig_tsne} shows the data visualization for top methods. We also visualize the local data of the six local models using t-SNE, plots are shown in Figures(~\ref{fig_tsne_LM1} --~\ref{fig_tsne_LM6}) and t-SNE for the Global model (our approach) is shown in~\ref{fig_tsne_GM}. We can see the improved grouping in the case of the Global model as compared to the local models. 
Figure~\ref{fig_tsne_Poin} shows t-SNE plots for Poincaré, it can be observed that the data points are scattered all over with no clear grouping, the reliability of these plots is further implied by the results shown in Table~\ref{Table_NN_Results}. The accuracy in the case of the Poincaré Embedding model is quite low which matches the poor grouping in the t-SNE plots. On the other hand, we can see the Autoencoder+NTK forms a clear grouping in the t-SNE plot(Figure ~\ref{fig_tsne_Auto_NTK}), this explicit grouping helps the classifier in achieving high accuracy which is proved by the results of Autoencoder+NTK based model shown in Table~\ref{Table_NN_Results}. Thus we can see clearly the t-SNE plots testify that the proposed approach leads to a better global model.

\section{Results And Discussion}\label{sec_ER}
In this section, we present the experimental results of Protein sequence classification using the proposed Dynamic Weighted Federated Learning (DWFL) and compare it with traditional protein classification methods including the Poincaré Embedding model (Feature Engineering),  Neural Network models including LSTM, GRU, CNN, Centralized Feed Forward NN, and Autoencoder + NTK. 
Table~\ref{Table_NN_Results} shows the outcomes of the classification tasks using various evaluation metrics (Accuracy, Precision, Recall, Weighted F1, Macro F1, and ROC AUC). We can observe from the results that DWFL (our method) outperforms all the baseline models except for Autoencoder + NTK, this observation is by all the evaluation metrics except for Train Time. Although Autoencoder + NTK achieves higher accuracy than our method it does not provide data privacy and it is a centralized method that requires data to be transmitted to a central server. 

\begin{figure}[h!]
\centering
\begin{subfigure}{.50\textwidth}
  \centering
  \includegraphics[scale=0.34]{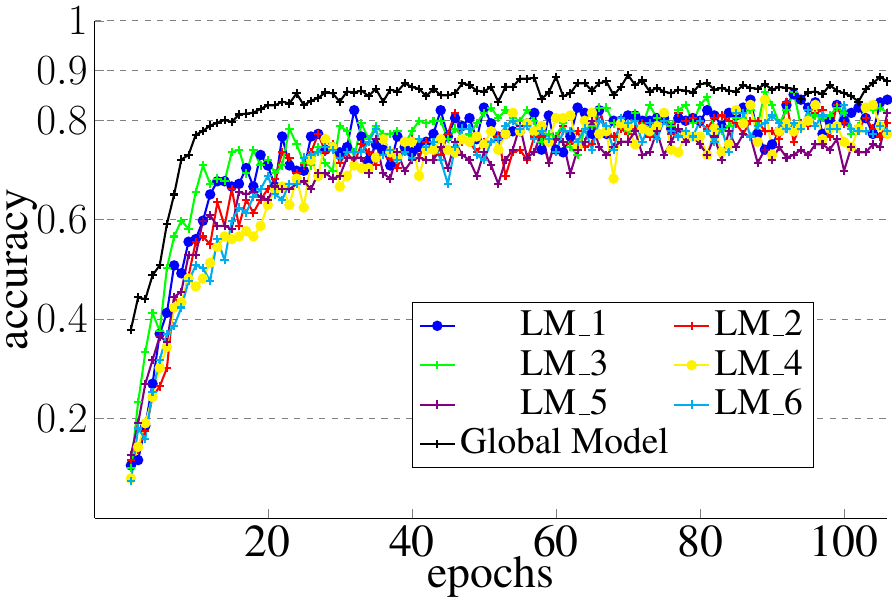}  
  \caption{Training Acc. - Local and Global Model}
  \label{fig_train_acc}
\end{subfigure}%
\begin{subfigure}{.50\textwidth}
  \centering
  \includegraphics[scale=0.34]{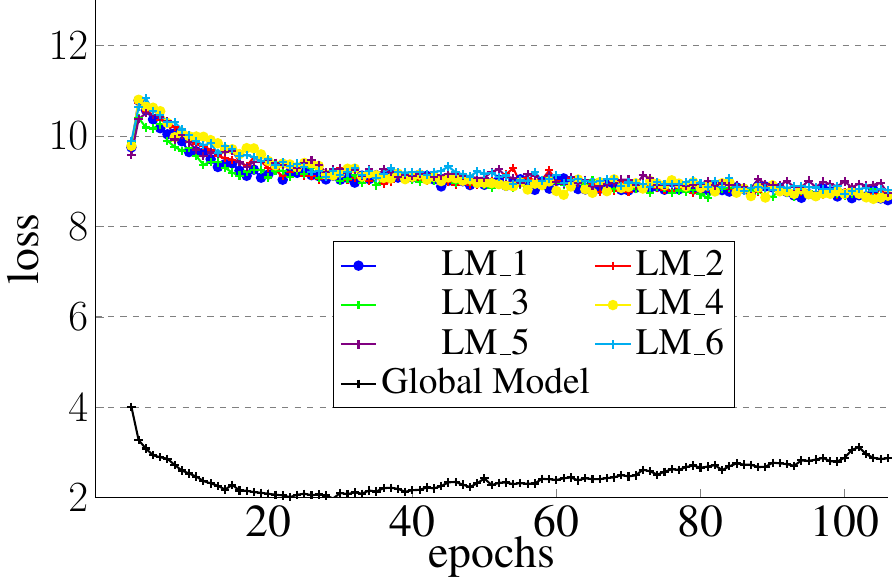}
  \caption{Training Loss - Local and Global Model}
  \label{fig_train_loss}
\end{subfigure}%
\caption{Training accuracy and loss of different local models and a Global model (DWFL) with increasing epochs (x-axis). The figure is best seen in color.
}
\label{fig_local_global_model_acc_loss_plots}
\end{figure}

Dynamic Weighted Federated Learning (DWFL) is a Federated Learning-based solution that improves predictive performance while maintaining privacy. We compare the proposed method with other Federated Learning approaches FLAvgWeight, FLMinWeight, and FLMaxWeight which perform simple weight aggregation instead of dynamic weight averaging. We can see from the results that our method DWFL outperforms the FLAvg in most evaluation metrics including Accuracy, Precision, Recall, and Weighted F1. This proves that our method DWFL is an enhanced Federated Learning approach that maintains privacy and improves performance. 
Training accuracy and loss of proposed DWFL are shown in Figure~\ref{fig_local_global_model_acc_loss_plots}. Figure~\ref{fig_train_acc} represents 
training accuracy with increasing epochs, and Figure~\ref{fig_train_loss} shows the training loss. It can be observed in Figure~\ref{fig_train_acc} that our Global model (represented by the black line) learns from the local models to achieve better accuracy. 
Additionally, the desirability of our global model's performance is further proved by observing Figure~\ref{fig_train_loss} in which the training loss of our Global model (represented by the black line) is lesser as compared to all the local models.

\begin{table}[h!]
    \centering
    \resizebox{0.85\textwidth}{!}{
    \begin{tabular}{p{1.4cm}cp{1.2cm}p{1.2cm}p{1.2cm}p{1.2cm}p{1.2cm}p{1.3cm}p{0.9cm}p{1.7cm}}
    \toprule
        \multirow{2}{*}{Category} & \multirow{2}{*}{Method} & \multirow{2}{*}{Classifier}  & \multirow{2}{*}{Acc. $\uparrow$} & \multirow{2}{*}{Prec. $\uparrow$} & \multirow{2}{*}{Recall $\uparrow$} & \multirow{2}{1.5cm}{F1 (Weig.) $\uparrow$} & \multirow{2}{1.5cm}{F1 (Macro) $\uparrow$} & \multirow{2}{1.2cm}{ROC AUC $\uparrow$} & Train Time (sec.) $\downarrow$
          \\
        \midrule \midrule
         \multirow{6}{1cm}{Feature Engineering} & \multirow{6}{1.5cm}{Poincaré Embedding} 
         & SVM & 0.3802 & 0.3246 & 0.3802 & 0.3194 & 0.1170 & 0.5237 & 0.8352	\\
        & & NB & 0.7189 & 0.7013 & 0.7189 & 0.7067 & 0.3170 & 0.6460 & \textbf{0.0041}   \\
        & & MLP & 0.6285 & 0.5149 & 0.6285 & 0.5548 & 0.1886 & 0.5772 & 1.7099  \\
        & & KNN & 0.5889 & 0.5592 & 0.5889 & 0.5548 & 0.2070 & 0.5801 & 0.0348  \\
        & & RF & \underline{0.7807} & 0.7432 & \underline{0.7807} & 0.7319 & 0.3264 & 0.6473 & 1.3720   \\
        & & LR & 0.3358 & 0.2713 & 0.3358 & 0.2686 & 0.0808 & 0.4978 & 0.0505   \\
        & & DT & 0.7576 & \underline{0.7564} & 0.7576 & \underline{0.7564} & \underline{0.3380} & \underline{0.6741} & 0.1060   \\

        \midrule
        \multirow{13}{1.2cm}{Central NN Baselines} & \multirow{7}{1.2cm}{Autoencoder + NTK} 
        & SVM & 0.8212 & 0.8119 & 0.8212 & 0.7992 & 0.5633 & 0.8109 & 0.1267	\\
        & & NB & 0.6660 & 0.7713 & 0.6660 & 0.6910 & 0.5987 & 0.8623 & \underline{0.0069}   \\
        & & MLP & 0.8554 & 0.8528 & 0.8554 & 0.8487 & 0.4347 & 0.7155 & 2.0283  \\
        & & KNN & 0.9071 & 0.9035 & 0.9071 & 0.9045 & 0.5071 & 0.7655 & 0.0385  \\
        & & RF & \textbf{0.9550} & \textbf{0.9560} & \textbf{0.9550} & \textbf{0.9546} & \textbf{0.8802} & \textbf{0.9470} & 0.5690   \\
        & & LR & 0.7933 & 0.7779 & 0.7933 & 0.7750 & 0.3669 & 0.6788 & 0.0371   \\
        & & DT & 0.9247 & 0.9272 & 0.9247 & 0.9253 & 0.7678 & 0.9104 & 0.0271   \\

        \cmidrule{2-10}
            & LSTM & - & 0.3829 & 0.1467 & 0.3829 & 0.2121 & 0.0503 & 0.5000 & 5068.3544 \\
            & GRU & - & 0.3105 & 0.0964 & 0.3105 & 0.1472 & 0.0474 & 0.5000 & 3694.2382 \\
            & CNN & - & 0.1708 &  0.0064 &  0.0802 & 0.0120 & 0.00149 & 0.5000 & \underline{423.5055} \\
            & Feed Forward NN & - & \underline{0.8021} & \underline{0.7003} & \underline{0.8021} & \underline{0.7310} & \underline{0.2999} & \underline{0.6404} & 657.0029 \\          
          \midrule
          \multirow{3}{1.4cm}{Federated Learning} 
          & FLAvgWeight & - & \underline{0.8745} & \underline{0.7879} & \underline{0.8745} & \underline{0.8245} & \underline{0.5205} & \underline{0.7529} & \underline{754.3194} \\
          & FLMinWeight & - & 0.6766 & 0.4879 & 0.6766 & 0.5581 & 0.1856 & 0.5830 & 860.5551 \\
          & FLMaxWeight & - & 0.7775 & 0.6207 & 0.7775 & 0.6885 & 0.2249 & 0.6085 & 784.1322 \\
          \midrule
          \multirow{3}{1.4cm}{Federated Learning (ours)} & DWFL-Avg & - & \textcolor{blue}{\underline{0.8913}} & \textcolor{blue}{\underline{0.8082}} & \textcolor{blue}{\underline{0.8913}} & \textcolor{blue}{\underline{0.8460}} & \textcolor{blue}{\underline{0.4763}} & \textcolor{blue}{\underline{0.7432}} & \textcolor{blue}{\underline{1063.0935}} \\
          & DWFL-Min & - & 0.8745 &  0.8911 & 0.8745 & 0.8300 & 0.3883 & 0.6854 & 1174.3684 \\
          & DWFL-Max & - & 0.8732 & 0.7958 &  0.8732 & 0.8269 & 0.3859 & 0.6847 & 1109.9863 \\
          \bottomrule
    \end{tabular}
    }
    \caption{ Results comparison of different neural network-based methods with proposed Federated Learning-based approaches. The best values for each model are underlined. Moreover, the overall best values are shown in bold. The best results for our proposed method are shown in blue color. 
    }
    \label{Table_NN_Results}
\end{table}

\section{Conclusion}\label{sec_conclusion}
In the context of analyzing protein sequence data, we propose a Dynamic Weighted Federated Learning (DWFL) solution to improve the predictive performance of biological sequence classification while maintaining data privacy and data availability. In the method we propose, the weights are changed by a dynamic component that evaluates the contribution of each local model's weight in light of its performance. We tested our method on a real-world dataset to demonstrate its efficacy and to show that it improves the classification performance compared to the baselines. 
In future work, we intend to use the proposed DWFL model on other molecular sequence datasets, we further intend to broaden the suggested strategy to encompass domains other than sequence data and assess the effectiveness of the strategy in those domains. We can also try applying this model in fields other than bioinformatics such as Energy informatics.

\bibliographystyle{splncs04}
\bibliography{8}

\clearpage

\end{document}